\documentclass[letterpaper, 10 pt, conference]{ieeeconf}  %

\IEEEoverridecommandlockouts                              %

\usepackage[colorinlistoftodos]{todonotes}
\usepackage[acronym]{glossaries}
\usepackage{algorithm}
\usepackage{algpseudocode}
\usepackage{soul}
\usepackage{multirow}
\usepackage{amssymb}
\usepackage{multicol,caption}
\usepackage{lipsum}
\usepackage{siunitx}
\usepackage{subcaption}
\usepackage{tabularx}
\usepackage{xcolor}
\usepackage{comment}
\usepackage{datetime}
\usepackage{enumerate}
\usepackage{booktabs}
\usepackage{flushend}
\usepackage{caption}

\newcommand{\second}{\text{ s}}

\newcommand{\Control}{\textnormal{C}}%
\newcommand{\Storyboard}{\textnormal{S}}%
\newcommand{\Video}{\textnormal{V}}%
\newcommand{\Text}{\textnormal{T}}
\newcommand{\G}{\textnormal{G}}
\newcommand{\Q}{\textnormal{Q}}
\newcommand{\score}{\textnormal{s}}

\newcommand{\question}{$question$}

\title{\LARGE \bf
`What did the Robot do in my Absence?' \\ Video Foundation Models to Enhance Intermittent Supervision 
}

\author{Kavindie Katuwandeniya$^{1}$, Leimin Tian$^{1}$, Dana Kuli\'c$^{2}$%
\thanks{$^{1}$CSIRO Robotics, Research way, Clayton, VIC, Australia. 
        {\tt\small Kavi.Katuwandeniya@csiro.au}}%
\thanks{$^{2}$Monash University, Clayton, VIC, Australia. D. Kuli{\'c} was supported by the ARC Future Fellowship (FT200100761).
{}
}
}

\begin{document}

\maketitle
\thispagestyle{empty}
\pagestyle{empty}

\begin{abstract}
This paper investigates the application of Video Foundation Models (ViFMs) for generating robot data summaries to enhance intermittent human supervision of robot teams. We propose a novel framework that produces both generic and query-driven summaries of long-duration robot vision data in three modalities: storyboards, short videos, and text. 
Through a user study involving $30$ participants, we evaluate the efficacy of these summary methods in allowing operators to accurately retrieve the observations and actions that occurred while the robot was operating without supervision over an extended duration (\qty{40}{\min}). Our findings reveal that query-driven summaries significantly improve retrieval accuracy compared to generic summaries or raw data, albeit with increased task duration. Storyboards are found to be the most effective presentation modality, especially for object-related queries. This work represents, to our knowledge, the first zero-shot application of ViFMs for generating multi-modal robot-to-human communication in intermittent supervision contexts, demonstrating both the promise and limitations of these models in human-robot interaction (HRI) scenarios.

\end{abstract}

\section{INTRODUCTION}

Advancements in artificial intelligence (AI) and robotics are leading to increased deployment of robots in challenging environments.  While current robots, operating in challenging environments are frequently teleoperated and require close human supervision, there is a drive towards increased robotic autonomy, enabling the robot to work for longer periods, such as for hours or even days, with no supervision. 

When an autonomous robot encounters exceptions or challenges in real-world deployment, it is common for the robot to request intervention from a human supervisor, shifting from full autonomy to partial autonomy or teleoperation, i.e.\ \emph{intermittent supervision}. 

To facilitate effective intermittent supervision, a robot needs to inform its human operator `what has been happening' while it has been operating autonomously, which can extend over long durations. Current approaches often involve requiring the human operator to remain attentive throughout the deployment~\cite{kottege2023heterogeneous} or 
manually examining raw data collected by the robot, such as video recordings from onboard cameras or robot status logs,  which is time-consuming, incurs high cognitive load, and requires expert knowledge to interpret~\cite{prewett2010managing}. 
This limits an expert operator's efficiency in supervising multiple robots and makes it challenging for novice operators to provide intermittent supervision. Therefore, we are motivated to investigate how data collected during autonomous robot operation can be distilled to facilitate efficient human supervision.

Recent advancements in Large Language Models (LLMs) for robotics have paved the way for a more natural human-robot interaction.  However, most existing research focuses on mapping natural language commands to a robot's plan and actions~\cite{lynch2023interactive}, 
or for a robot to generate conversational responses in social interactions~\cite{grassi2022knowledge}. %
Using LLMs to generate robot-to-human communication to aid human comprehension of robot status and intervention in functional tasks remains largely unexplored, with only a single study~\cite{dechant2023learning} exploring the use of a custom-trained LLM for generic and query-driven robot action summarisation. 

\begin{figure}
    \centering    \includegraphics[width=0.45\textwidth]{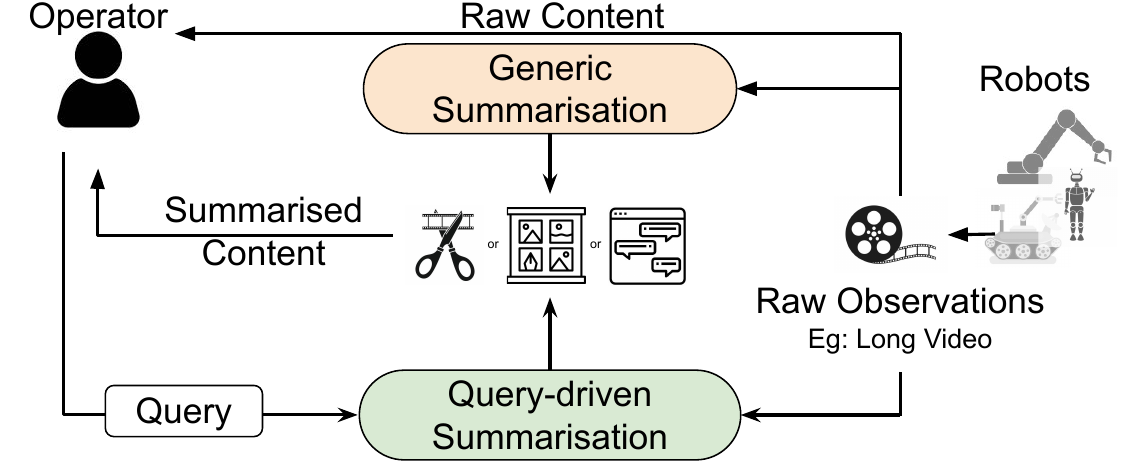}
    \caption{Diagram of the proposed robot summary generation system. The system generates generic or query-driven summaries of long egocentric robot videos in the form of storyboards, short videos, or text to help a user review the robots' autonomous history.}
    \vspace{-5mm}
    \label{fig:mainfig}
\end{figure}

We propose to employ Foundation Models to summarise robot data and improve intermittent supervision in HRI, as shown in Figure~\ref{fig:mainfig}. In particular, we propose to use pre-trained Video Foundation Models (ViFMs) to summarise long (\qty{40}{\min}) egocentric robot videos to assist operators to rapidly identify object occurrences and temporal events in a multi-robot search-and-rescue scenarios~\cite{kottege2023heterogeneous}. 

There are two main challenges in adopting ViFMs for generating robot summaries for humans: 
\begin{enumerate}
\item Summaries generated by ViFMs may not capture relevant information for a specific task. Thus, a user may need to explicitly query the model for targeted answers. 
\item The generated summary may be inaccurate. Thus, the modality in which the summary is presented may influence a user's ability to verify the summary and calibrate their trust towards the information provided.
\end{enumerate}
To investigate ViFMs for robot summary generation, we develop a framework that provides a user with either generic or query-driven summaries of robot videos in the following output modalities: a `storyboard' of key images (static summarisation), a short summary `video' (dynamic summarisation), and a language-based `text' summary. We evaluate the proposed framework in a user study (n = 30) to compare participants' task performance and subjective experience using these different summary generation methods.

To the best of our knowledge, this is the first work which incorporates ViFMs in a zero-shot setting to generate summary communication in multiple natural modalities from a robot to a human for intermittent supervision. Our work illustrates the benefits and limitations of such models, as well as the transferability of ViFMs trained with general data for a specific HRI task scenario.

\section{Related Work}

We first review the related work in video summarisation, an active topic in the machine learning literature~\cite{meena2023review}.  Next, we review the literature on ViFMs and their application to video summarisation.  Finally, we consider applications in human-robot communication.

\subsection{Video Summarisation}
Videos are described by segments containing frames~\cite{meena2023review}, displayed at a specific frequency to create motion~\cite{akhare2022query}. 
Initial works in video summarisation considered video data without connection to natural language.  
Video summarisation distills content while preserving key information~\cite{akhare2022query}. 
There are two main approaches to video summarisation: static and dynamic~\cite{meena2023review}. 
Static (key-frame) summarisation selects representative frames to create a storyboard, while dynamic summarisation (video-skimming) identifies and chronologically arranges key video segments. Static summaries are more compact, whereas dynamic summaries tend to be more informative~\cite{meena2023review}.  
SumMe~\cite{gygli2014creating} and TVSum~\cite{song2015tvsum} are two benchmark datasets commonly used in video summarisation, both provide human-annotated importance scores for frames. 
We will refer to this line of work as `generic video summarisation'. 

Transformer-based video summarisation methods~\cite{zhao2022hierarchical, hsu2023video} excel in extracting global dependencies and multihop relationships between video frames. 
Spatiotemporal vision transformer (STVT) model~\cite{hsu2023video} achieved SOTA performance on the SumMe and TVSum datasets by training the model using both inter-frame and intra-frame information. 

To address the user subjectivity in the summaries, Sharghi~et~al~\cite{sharghi2017query} introduce `query-focused
video summarisation' where the user preferences were introduced to the summarisation process in the form of text queries. 
However, these queries were limited in scope and vocabulary. 
With the advancements in LLMs and later in
multi-modal foundation models (ViFMs are a subset of
this), open-ended, natural language queries became possible, enabling more sophisticated and contextually relevant video summarisation. 
We refer to these summaries as `query-driven' summaries.

\subsection{ViFMs for Video Summarisation}
ViFMs learn a general-purpose representation for various video understanding tasks, by leveraging large-scale datasets. 
A detailed survey of ViFMs can be found in~\cite{madan2024foundation}. 
These models are typically evaluated on diverse benchmark datasets, e.g., MVBench~\cite{li2023mvbench} that includes various spatio-temporal tasks. 

Video summarisation is not a common task for ViFM. Therefore, we explore video understanding tasks capable of generating video skims, storyboards, or textual outputs based on a video and a natural language query. It is important to note that traditional video summaries do not typically include text as a modality~\cite{meena2023review}.

Madan~et~al~\cite{madan2024foundation} classify various video understanding tasks into $3$ broad categories: video content understanding, descriptive understanding tasks, and video content generation and manipulation. Our interest lies in the first $2$ with the first generating visual content and the second, text. 
Video content understanding tasks are again categorised into three levels: abstract understanding (e.g., classification, retrieval), temporal understanding (e.g., action localization), and spatio-temporal understanding (e.g., object tracking, segmentation). Descriptive understanding tasks, such as video question answering (VQA) and captioning, focus on understanding of the textual description of the video content. 

In terms of generating summaries as visual outputs, retrieval involves finding video segments containing specific actions, objects, or scenes, going beyond just objects or just actions, and is most closely related to our work.  
LanguageBind~\cite{zhu2023languagebind} achieves SOTA performance in zero-shot video-text retrieval, using language as a binding modality to establish semantic connections between textual descriptions and corresponding visual content. 

While both tasks under descriptive video understanding can generate textual summaries, those models that excel in VQA are preferred for their ability to provide summaries for subjective, creative, and logical questions. VideoChat2~\cite{li2023mvbench} excelled in challenging temporal video understanding tasks on the MVBench~\cite{li2023mvbench} benchmark by outperforming SOTA ViFMs by over $15\%$.

\subsection{Foundation Models for Robot Communication}
Firoozi~et~al~\cite{firoozi2023foundation}  provided an extensive survey on how foundation models improve robot capabilities such as robot learning or perception. 
However, the use of foundations models for robots communicating with humans for intermittent supervision, remains a crucial, yet underexplored, aspect. 

Tellex~et~al.~\cite{tellex2020robots} reviewed research on natural language use in robotics, finding that most research focuses on robots following human instructions~\cite{lynch2023interactive}. 
There is limited research on robot-to-human communication or two-way communication, which is what this study addresses. 
Specifically, we investigate unprompted generic summarisation (robot-to-human communication), and interactive `query-driven summarisation' (two-way communication). 
Dechant~et~al.~\cite{dechant2023learning} were the first to use an LLM for generic and query-driven robot action summarisation. 
They trained a single model to both summarise and answer questions. 
Our work differs by proposing a general framework that allows the integration of any existing ViFM  without dataset-specific finetuning. 
While they used a static validation set, we emphasise human-in-the-loop evaluation and include visual modalities for added clarity and context, not just text. 
Moreover, the proposed framework supports intermittent supervision by handling longer videos, and going beyond summarising robot actions to include world observations and objects.

\begin{figure}[t]
    \centering
    \includegraphics[width=0.48\textwidth]{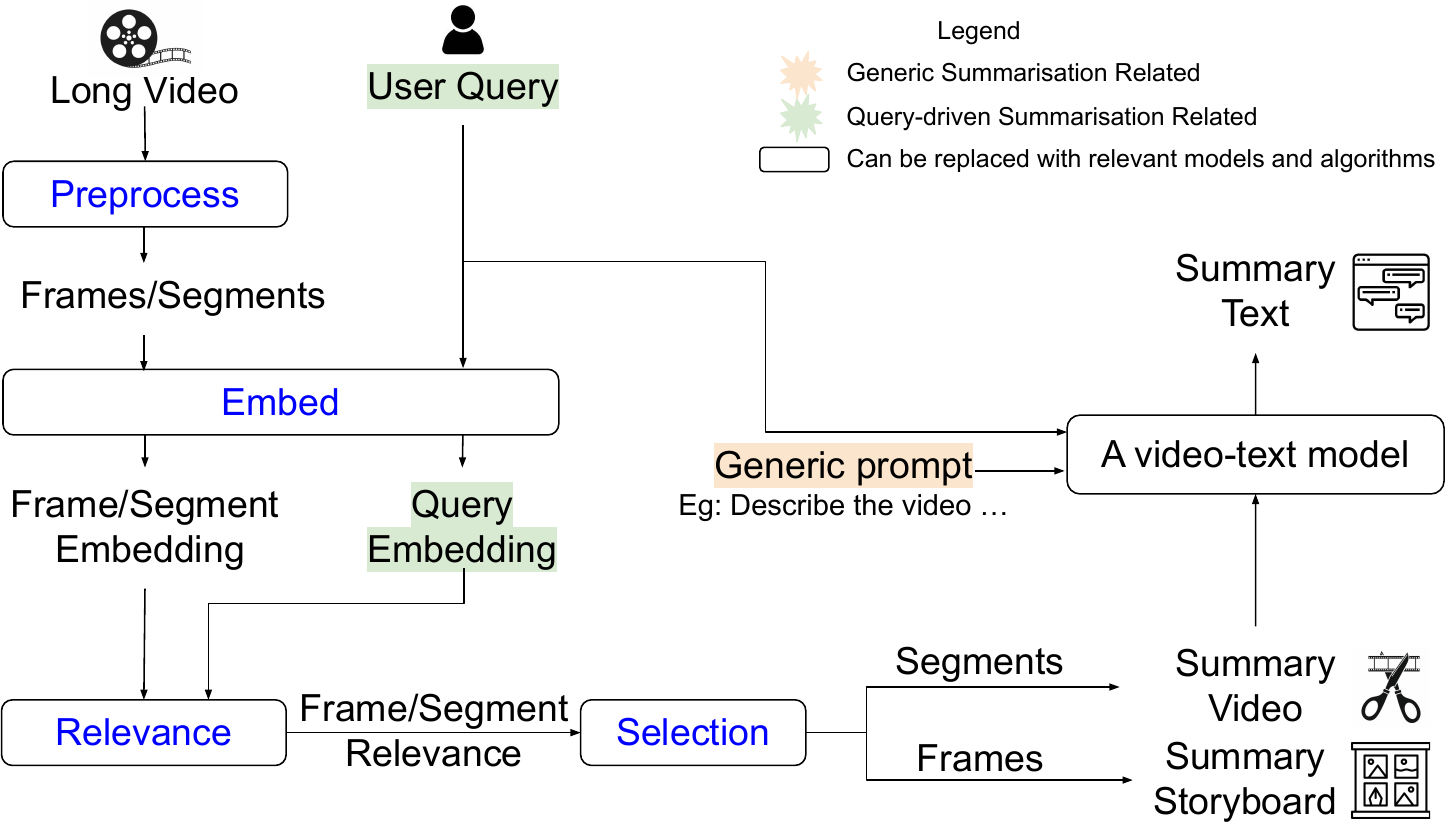}
    \caption{Proposed framework for generating generic and query-driven summaries in the form of storyboard, video or text. There are $4$ main steps: \textcolor{blue}{preprocess, embed, relevance}, and \textcolor{blue}{selection}. The models and algorithms used for each step can be replaced with similar models and algorithms.}
    \vspace{-5mm}
    \label{fig:system_arch}
\end{figure}

\section{FRAMEWORK DESIGN}

There are two main challenges in adopting ViFMs for generating robot summaries for humans. 
First, how should a summary be generated?  A generic summary generation process may require less user effort, but the generic summary may not capture relevant information for a specific task.  On the other hand, explicit user queries may provide more relevant summaries, but may require additional user expertise, time and effort.  
Second, how should the summarised information be presented to the user?  
Summaries can be presented in different mediums such as text, images or videos.   
The presentation modality may influence  the user's ability to verify the information and appropriately calibrate their trust in the summary.

To investigate ViFMs for robot summary generation, we developed a framework to provide users with generic or query-driven summaries in the form of a short summary video (video-skims): \Video, a storyboard of key images: \Storyboard, or a language-based text summary: \Text. 
In addition, a user study is conducted to compare participants' task performance and subjective experience using these different summary generation methods. 

The proposed framework is shown in Fig.~\ref{fig:system_arch}. 
There are four main steps: `preprocessing', `embedding', `relevance generation', and `selection'. 
The specifics for the two pipelines are given in Tab.~\ref{tab:processes} with further explanation under Sec.~\ref{subsec:generic} and Sec.~\ref{subsec:query}.  
In both pipelines, the text summary is based on the generated summary video instead of the original video to mitigate the model's memory constraints (videochat2~\cite{li2023mvbench}).

\begin{table*}[thpb]
\caption{Detailed description of each process in the proposed framework shown in Fig.~\ref{fig:system_arch}. }
\vspace{-2mm}
\label{tab:processes}
\begin{tabular}{|c|cc|cc|}
\hline
\multirow{2}{*}{Process} &
  \multicolumn{2}{c|}{Generic \G} &
  \multicolumn{2}{c|}{Query \Q} \\ \cline{2-5} 
 &
  \multicolumn{1}{c|}{Storyboard \Storyboard} &
  Video \Video \& Text \Text &
  \multicolumn{1}{c|}{Storyboard \Storyboard} &
  Video \Video \& Text \Text \\ \hline
Preprocessing &
  \multicolumn{2}{c|}{Frames at original frame rate (\qty{15}{fps})} &
  \multicolumn{1}{c|}{Frames at a  lower frame rate  (\qty{1}{fps})} &
  \begin{tabular}[c]{@{}c@{}}Frames at a lower rate (\qty{1}{fps})\\ Fixed length segments\\  ($d=8\second$ as per~\cite{zhu2023languagebind})\end{tabular} \\ \hline
Embedding &
  \multicolumn{2}{c|}{ResNet18~\cite{he2016deep} ($512$ dimensional vector)} &
  \multicolumn{2}{c|}{LanguageBind~\cite{zhu2023languagebind} ($768$ dimensional vector)} \\ \hline
Relevance &
  \multicolumn{2}{c|}{STVT~\cite{hsu2023video}} &
  \multicolumn{2}{c|}{Dot Product} \\ \hline
Selection &
  \multicolumn{1}{c|}{\begin{tabular}[c]{@{}c@{}}PCA ($L=100$)\\ Greedy ($\delta=0.5, M=24$)\end{tabular}} &
  \begin{tabular}[c]{@{}c@{}}PCA ($L=100$)\\ KTS ($D=1\second$)~\cite{potapov2014category}\\ Knapsack ($K=15\%$)~\cite{hsu2023video}\end{tabular} &
  \multicolumn{1}{c|}{Greedy ($\delta=0.5, m=4$)} &
  Top~$k$ ($k=6$) \\ \hline
\end{tabular}
\end{table*}

\subsection{Generic Summary  Pipeline}
\label{subsec:generic}

A generic summary is generated without considering any input from the user.  
In the preprocessing step, the original long video is taken at its original frame rate. %
Then, each frame is encoded with ResNet18~\cite{he2016deep} and fed into the relevance generation process to generate frame-level importance.  
Frame-level importance is generated by STVT~\cite{hsu2023video}. 
By combining the temporal inter-frame correlations 
and the spatial intra-frame attention within frames, STVT captures important temporal and spatial content. The model was trained to learn the frame importance for each frame from human-created summaries on the TVSum dataset.  
The choice for embedding depends on the model used to generate the relevance scores. 
While the videos used to train STVT range from \numrange{2}{10} minutes, the model is scalable for longer videos as the video is considered a collection of individual frames.   

During the selection process, important frames are selected to create a generic storyboard while important segments are selected to create a generic video skim. 
To generate the storyboard, a greedy approach is followed, where the frames selected ($\# M$) %
have a high frame level importance but low semantic similarity to those that are already selected. Semantic similarity is based on the top $L$ Principal Component Analysis (PCA) features of the ResNet18 features of each frame. A threshold value of $\delta$ is used to determine  if the frame is similar to any previously selected frames.  
The images are then presented to the user in a storyboard chronologically, along with the relevant timestamp. The individual relevance scores are not shown.  

To generate a summary video, first, the frame-level sequence is transformed to a segment-level sequence. 
A widely used technique for this is Kernel-based Temporal Segmentation (KTS)~\cite{potapov2014category}. 
We use a cosine kernel change point detection algorithm, where the number of change points is not specified in the optimization algorithm.    
KTS receives frame level features, generated in the same way as for the storyboard modality.      
We constrain each selected segment to a minimum duration of~$D$ to prevent the effect of simply producing a video played at a higher speed. 
The change points define the segment boundaries, and the segment's relevance score is calculated as the summation of frame-level scores within the segment. 
The \{0/1\}~knapsack algorithm~\cite{hsu2023video} is used to select the segments that have the highest importance score. 
The duration of the knapsack selection is specified as a percentage~$K\%<100\%$ of the original video duration, generating a generic summary video. %

A generic text description is produced by Videochat2 based on the generic video summary. 
The model receives a generic query: it is given context about the robot's environment and task, and asked to describe the video in detail.

\subsection{Query-Driven Summary  Pipeline}
\label{subsec:query}

The main difference between the generic summary and the query-driven summary is the additional input of a user query, provided as a text input. %
The users are free to query however they wish to understand what the robot has been doing in their absence, going beyond querying from a limited set of options adopted by previous research~\cite{wu2022intentvizor}. 

In the preprocessing step, the long video is either broken down into frames (for storyboard) or segments (for video and text). 
The original video can be read at its original frame rate or a lower frame rate, depending on the availability of computational resources. 
The duration of a segment~$d$ depends on the requirements of the model used to embed the segment. 

Both the text-based user query and the frames/segments are embedded using LanguageBind~\cite{zhu2023languagebind}. 
To generate relevance scores, the dot product between the query embedding and frame/segment-level embedding is used. 
It is important that both the query embedding and frame/segment embedding are in the same embedding space for the dot product to generate a valid similarity score. 

Selecting the frames ($\# m$) for the query-driven storyboard follows the same greedy approach from the generic pipeline, except the embeddings are based on LanguageBind.  
Similarly, the images are then presented chronologically, irrespective of their relevance score. 

The top~$k$ video segments are selected and combined chronologically, to generate the query-driven summary video. 
Similar to the generic text summary description, the query-driven summary video is fed to the videochat2~model, along with the user query, to generate a query-driven summary text.

\begin{figure}
  \centering
  \includegraphics[scale=0.4]{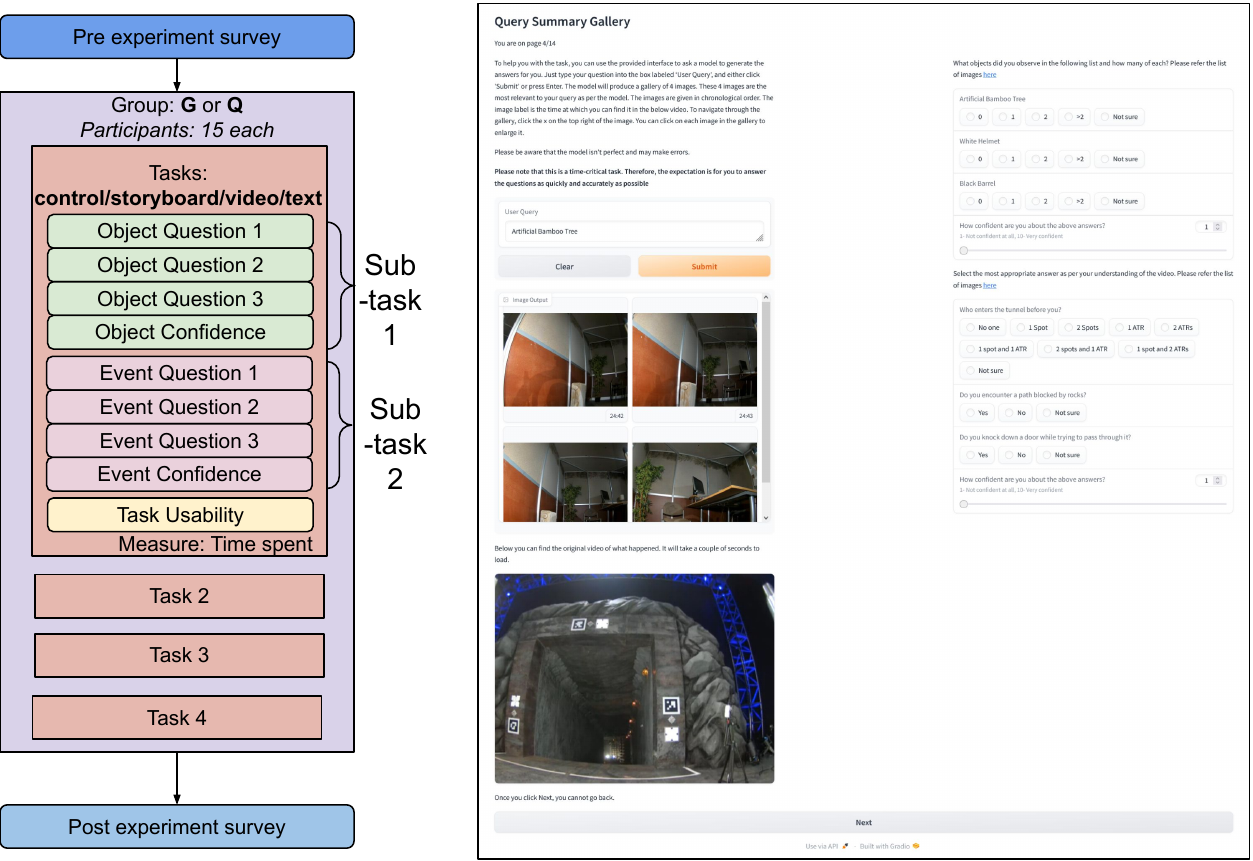} %
  \caption{(\textit{left}): Study procedure and (sub-)tasks. (\textit{right}): User study interface for query-driven storyboard summary. When the user enters a query into the `User Query' box, selected images are displayed below the box based on the raw video given at bottom left. 
  Questions are given on the right side.}
  \vspace{-5mm}
  \label{fig:interface}
\end{figure}

\subsection{User Study Design}

With this framework in place to generate summaries to inform a user about what the robot has been doing in their absence, we designed a user study to evaluate whether users could benefit from such a summarisation framework.  
We consider a search and rescue mission undertaken by a fleet of robots in a complex underground environment (tunnel, urban, and cave) where they explore the previously unknown environment, searching for specific objects~\cite{kottege2023heterogeneous}. 
After extended autonomous exploration, a human operator is asked to identify key objects and events encountered by the robots during the operator's absence by either reviewing the raw video data, or summaries of the data generated by the proposed framework.

The user study is developed as a mixed-model design to investigate the effects of using summaries by ViFMs on task performance and usability. 
The user study protocols have been reviewed and approved by the CSIRO research ethics committee. 
A total of \num{30}~participants were recruited for the study. The flow of the study is shown in Fig.~\ref{fig:interface} \textit{(left)}. A pre-study survey was conducted to understand the 
familiarity of the participants with robotics and video summarisation models, and their trust of such models.

Participants were randomly assigned to either the generic summarisation pipeline~$\G$ or the query-driven summarisation pipeline~$\Q$. 
We will refer to this as `GvsQ'$\in\{\G,\Q\}$. 
Both groups also experienced the control \Control\ condition, which consisted of only the raw video without any AI-generated summaries. 
Users were allowed to watch the long raw video at \qty{2}{\times}, \qty{4}{\times} or \qty{8}{\times} speeds since this is a common way to review long videos. 
A within-subject design was used to compare the summarisation modalities: \Video, \Storyboard, and \Text. 
Here the participants were provided with both the raw video and the summaries, where they were free to play or ignore the full-length video with no speed control. 
Thus, each participant was given four intermittent supervision tasks (\Control, \Video, \Storyboard, \Text) in a randomised order, where each task was to answer a set of questions based on front-egocentric video feed of the robots (see Sec.~\ref{subsec:dataset}). 
Fig.~\ref{fig:interface} \textit{(right)} illustrates the user interface for the \Storyboard\ modality under the \Q\ pipeline.

Four different videos (and video-related questions) were used for the four tasks, with the same set of videos and questions used for all participants under \G\ and \Q\  conditions. 
The questions were designed with varying difficulty to assess the user's ability to identify relevant information in the video content with or without the aid of different types of summaries. 
Each task contains two sub-tasks with different types of questions: object-related and event-related, i.e., question types \question. 
Sec.~\ref{subsec:questions} describes the tasks and questions in more detail. 
The users were asked to answer the questions as accurately and efficiently as possible, and were reminded the tasks were time-critical.  
Participants' responses to the questions, their confidence, task completion times, and usability preference~\cite{brooke1996usability} were recorded for each task.

We examine the following hypotheses in this user study: 
\begin{itemize}
    \item \textbf{H1}: Distilling data improves: \textbf{H1a} accuracy, and \textbf{H1b} time,
    for retrieving relevant information as opposed to providing raw data 
    (task score of \G/\Q $>$\Control\ and task time of \G/\Q $<$\Control). 
    \item \textbf{H2}: The magnitude of the improvement in: \textbf{H2a} accuracy, and \textbf{H2b} time, depends on the summary type~(\G/\Q) and modality~(\Video/\Storyboard/\Text). 
\end{itemize}

\subsection{Implementation}

\begin{figure}
 \centering
 \begin{subfigure}[h]{0.20\textwidth}
     \centering
     \includegraphics[width=\textwidth]{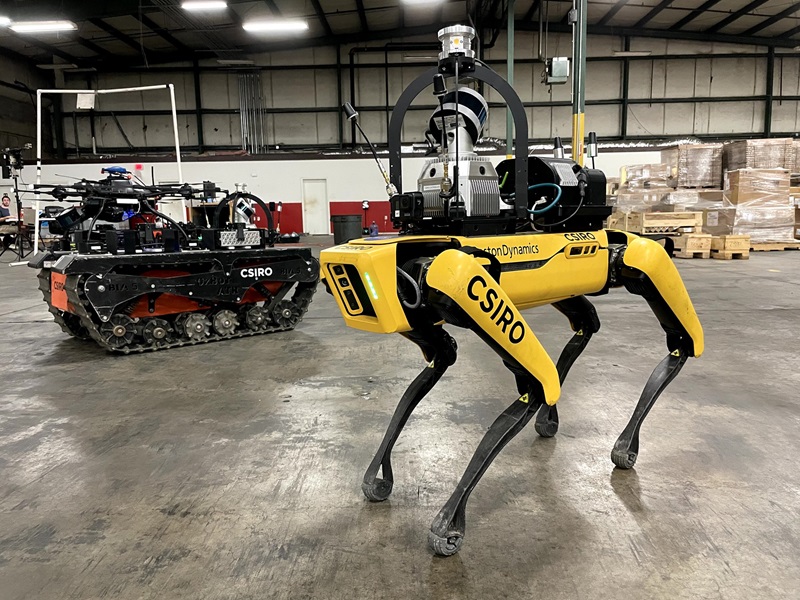}
 \end{subfigure}
 \hfill
 \begin{subfigure}[h]{0.20\textwidth}
     \centering
     \includegraphics[width=\textwidth]{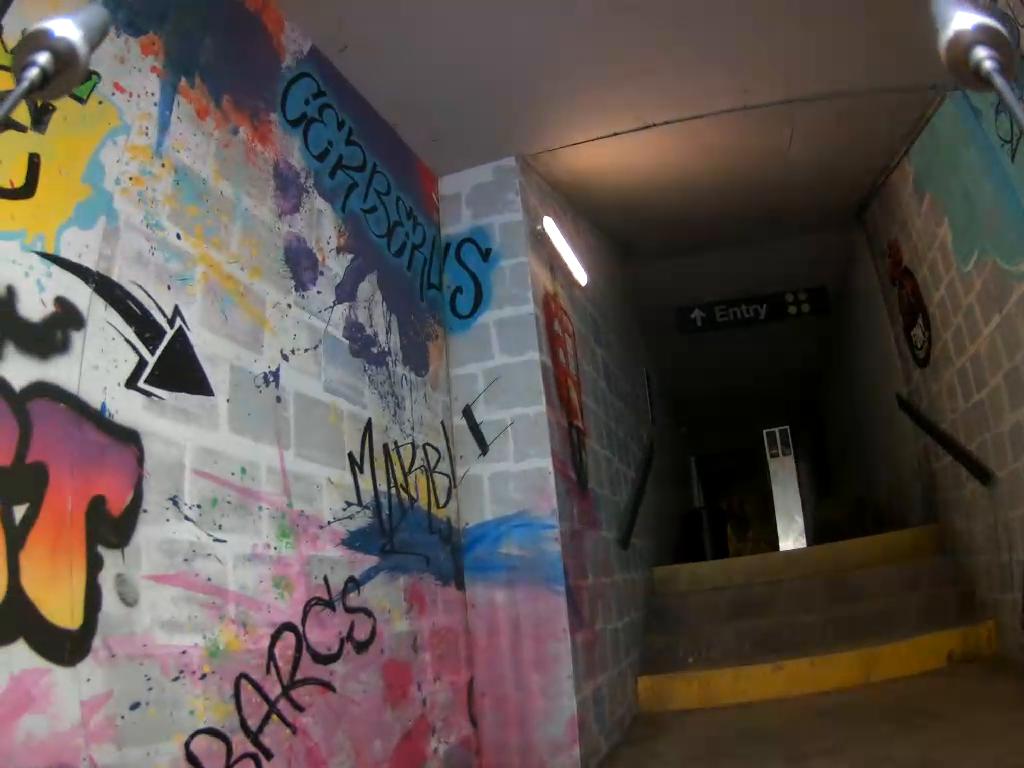}
 \end{subfigure}
    \caption{Example image \textit{(right)} from the front egocentric video feed from a fleet of robots \textit{(left)} deployed for an underground search and rescue mission~\cite{kottege2023heterogeneous}}
    \vspace{-5mm}
    \label{fig:dataset}
\end{figure}

\subsubsection{Dataset}
\label{subsec:dataset}
The videos (RGB data stream) used for the study were collected from a fleet of robots exploring complex environments, such as tunnels and urban undergrounds, completing a search and rescue mission~\cite{kottege2023heterogeneous}. 
Front-facing egocentric video of two robots are used: quadruped robot Spot and All-terrain Robot (ATR) shown in Fig.~\ref{fig:dataset} \textit{(left)}.  
The length of each video is \qty{40}{\min} read at \qty{15}{fps}.  
An example frame from the videos are shown in Fig.~\ref{fig:dataset} \textit{(right)}.

\subsubsection{Intermittent Supervision Tasks}
\label{subsec:questions} 
Each task (\Control, \Video, \Storyboard, \Text) includes \num{6} unique questions related to its corresponding video, distinct from the questions in other tasks. This design minimizes the potential influence of earlier tasks on later ones by preventing users from memorizing and anticipating similar questions. 
Altogether, \num{24} questions were designed with varying difficulty levels, inspired by the rescue mission~\cite{kottege2023heterogeneous} and MVBench~\cite{li2023mvbench}.

The \num{6} questions are broken down into two sub-tasks/question types. 
Questions \numrange{1}{3} were about the existence of objects of interest and their corresponding counts, aligning with rescue mission objectives. 
Questions \numrange{4}{6} were about events or actions requiring temporal understanding inspired by the temporal task definitions of MVBench, including `action sequence', `moving direction' and `state transitions'.
All questions are presented as Multiple Choice Questions with  `Not sure' included as an option.

\subsubsection{Computational Considerations}
Generic summaries were pre-generated while query-driven summaries were generated in real-time with the backend models running in an H100 GPU. 
As such, the computational requirements of the two pipelines were different. 
The models chosen under each pipeline also contribute to the difference in computational requirements.  
None of the models used for the implementation were finetuned for the test data. 

In the `preprocessing' stage, the \Q\ pipeline used a lower frame rate to manage the memory requirements when running in real-time specifically with ViFMs: LanguageBind and videochat2 simultaneously loaded to memory.

\subsubsection{Equalising information quality across conditions}
One crucial aspect of the study was to ensure a similar amount of information is provided in each modality across the two summary types. 

The threshold $\%$ for generating the generic video for the knapsack algorithm was selected to be $K=15\%$, similar to STVT. This resulted in a video of approximately %
\qty{6}{\min}. 
Under \Q, assuming users would enter at least \qty{6}{queries} (for the \qty{6}{questions}), the top-$k$ algorithm was set with $k=6$. This produced a summary with a duration of \qty{4}{\min} \qty{48}{\second}, 
resulting in nearly equal-length summary videos for both \G\ and \Q. 

In the query storyboard we chose $m=4$ images dependent on the interface design, allowing the user to visualise all images at the same time without scrolling. 
With the same assumption of  $6$ queries from a user, for the generic storyboard, we selected $M=24$ images in a scrollable interface.

To generate the text summary, in both pipelines videochat2~\cite{li2023mvbench} receives the generated video summary 
from which only \qty{100}{frames} are considered due to 
resource limitations. 
By controlling the length penalty, the length of the summary was made compatible in both pipelines: approximately six sentences for the \G\ summary and one sentence for each query in the \Q\ summary.

\section{RESULTS ANALYSIS AND DISCUSSION}
{ %
\sisetup{
    round-mode=figures,
    round-precision=3,
    round-pad=false,
}%

The analysis investigates the two hypotheses \textbf{H1} (summary brings benefits) and \textbf{H2} (summary type and modality matters) respectively. 
The first considers `summary type' (\G, \Q, `No Summary'=\Control), while the second excludes \Control\ data to compare modalities in \G vs\Q\ conditions.

Dependant variables are the score (\score) a user receives for each answer they provide and the time ($t$) they spend on each task. 
Scores are binary ($0$ for incorrect, $1$ for correct), with only one correct answer per question: any other answer, including `Not sure' is marked as incorrect. 
Time was recorded at the task-level.

When fitting models to explain the relationship between dependant and independent variables, 
the fitting formula 
is iteratively refined  by initially including all possible interactions (two-way, and three-way where relevant), then progressively eliminating insignificant ones until only statistically significant interactions remain. 
Score (binary) is modelled using logistic regression, while time (continuous), with ordinary least squares (OLS) regression. 
Resulting coefficients and the corresponding p-values are given in Tab~\ref{tab:regression}. 

\subsection{Importance of Distilling Information} 
We first analyse (\textbf{H1}): distilling information improves accuracy (\textbf{H1a}) and efficiency (\textbf{H1b}) as opposed to providing raw data. 
The independent variables consist of the summary type (\G, \Q, \Control) and, where relevant, question type \question\ which is a categorical variable with values `object' and `event'.  
The baseline model for \score\ is $summary[\Control]$ and $question[object]$, and for $t$, $summary[\Control]$  (only task-level data available).

\paragraph{Accuracy}
\label{subsubsub:first_accuracy}

Allowing users to query for specific information significantly improves accuracy (coef. = \num{0.549}, $p=$\num{0.005}) compared to providing raw information alone. Interestingly, no statistically significant relationship was observed with \G\ summarisation. These findings suggest that the effectiveness of information distillation in enhancing user accuracy is contingent upon the interactive nature of the summaries, partially supporting the hypothesis \textbf{H1a}: score of $\G/\Q > \Control$, where only score of $\G>\Control$. 

We also found a significant improvement in accuracy for event-related questions compared to object-related questions (coef. \num{0.618},  $p<0.001$).  
This indicates that the users were able to answer the event-related questions more accurately than the object-related questions. 
This observation falls in line with the videochat2 results~\cite{li2023mvbench} where the authors hypothesise that the current ViFMs have difficulty generalising to localization and counting tasks in the absence of related tuning data.

\paragraph{Efficiency}

\begin{table*}[]
\caption{Coefficients of regression models for user score $\score=\{0,1\}$ and time~$t$.}
\vspace{-2mm}
\label{tab:regression}
\begin{tabular}{@{}rlSSSS@{}}
\toprule
    &                       &\multicolumn{2}{c}{Score}  &\multicolumn{2}{c}{Time}       \\
                            \cmidrule(lr){3-4}          \cmidrule(lr){5-6}
Analysis    
    & Variable              &{Coefficient}&{p-value}    &{Coefficient}&{p-value}        \\
\midrule
\multirow{4}{*}{First}
    & Intercept             & -0.5839 & 0.001~**        & 473.6732  & 3.882052e-23~***  \\
    &$summary[\G]$          & -0.3010 & 0.130           & -28.9768  & 0.557             \\
    &$summary[\Q]$          & 0.5494  & 0.005~**        & 156.3481  & 0.002~**          \\
    &$question[event]$      & 0.6184  & 0.000060~***    &           &                   \\
\midrule
\multirow{7}{*}{Second}
    & Intercept             & -0.9008 & 0.0002~***      & 552.2573  & 2.161442e-20~***  \\
    &$\G vs\Q[\Q]$          & 0.9453  & 6.488738e-07~***& 185.3249  & 9.930407e-05~***  \\
    &$modality[\Text]$      & -0.9546 & 0.005~**        & -198.0177 & 6.029031e-04~***  \\
    &$modality[\Storyboard]$& 0.9054  & 0.004~**        & -124.6649 & 0.028~*           \\
    &$question[event]$      & 1.4922  & 6.183559e-06~***&           &                   \\
    &$modality[\Text]question[event]$
                            & -0.3452 & 0.466           &           &                   \\
    &$modality[\Storyboard]question[event]$
                            & -2.4466 & 8.828819e-08~***&           &                   \\
\bottomrule
\addlinespace
\multicolumn{6}{c}{Significance Codes: $0 < $ *** $ < 0.001 < $ ** $ < 0.01 < $ * $ < 0.05$} \\ 
\end{tabular}
\end{table*}

The results from the OLS model do not support the hypothesis \textbf{H1b}. 
While the use of \G\ summarisation showed a slight reduction in retrieval time (coef. \num{-28.9768}), this effect was not statistically significant ($p = 0.557$). 
Query-driven summarisation \Q\ was associated with a significant increase in retrieval time (coef. \num{156.348},  $p =$\num{0.002}). 
In the Q condition, the summaries are not readily available, as such there is a latency between querying and having a summary to inspect. 
This finding suggests that while query-based systems may improve accuracy, they may also introduce a time cost.

\subsection{Importance of GvsQ and Modality}
We next test the hypotheses \textbf{H2}: the magnitude of the improvement in accuracy (\textbf{H2a}) and time (\textbf{H2b})  depends on the summary type GvsQ and presentation modality~(\Video/\Storyboard/\Text). 
The independent variables for each model consist of the \G vs\Q\ and modality (and question type where relevant) which are all categorical.

\paragraph{Accuracy}
The baseline model for the analysis is $GvsQ[\G]$,  $modality[\Video]$ and $question[object]$. 
The results (Tab.~\ref{tab:regression}) support the hypothesis \textbf{H2a}. 
\Q\ summaries showed a substantial positive effect (coef. \num{0.9453}, $p< 0.001$) compared to the \G\ summaries, emphasising that allowing users to actively shape the information they receive leads to better accuracy than merely providing them generic summaries. 

Modality also played a crucial role, with \Text\ presentations significantly decreasing performance (coef. $\num{-0.9546}$, $p=0.005$) relative to \Video, while \Storyboard\ presentations showed a positive effect (coef. $\num{0.9054}$, $p = 0.004$). 
These findings suggest that the choice of modality can significantly influence task accuracy, with \Storyboard\ potentially offering advantages over both \Video\ and \Text\ in certain contexts (type of question matters - see below). 
A possible cause of the significantly decreased accuracy in \Text\ may be the lack of visual information for the user to verify the summary.  

The type of question demonstrated a significant main effect (\question[$event$], coef. $\num{1.492}$, $p< 0.001$) as with the first analysis. 
However, the interaction effects between modality and question type revealed complex patterns. While the interaction between \Text\ and event questions was not significant, the interaction between \Storyboard\ and event questions showed a strong negative effect (coef. $\num{-2.4466}$,  $p< 0.001$). 
This suggests that while storyboards generally improved performance, they were less effective for event-related questions compared to object-related questions: unlike a video, static images cannot show the unfolding of an event.

\paragraph{Efficiency} 
The baseline model for examining the time duration needed to complete the task, $t$,  is $GvsQ[\G]$ and $modality[\Video]$. 
\Q\ summaries showed a significant increase in time compared to the \G\ summaries (coef. $\num{185.3249}$,  $p< 0.001$), re-emphasising the time cost of interactive models. 

Regarding interaction modality, both \Text\ and \Storyboard\ presentations showed significant reductions in time compared to the \Video\ (coef. $\num{-198.0177}$, $p<0.001$; coef. $\num{-124.6649}$, $p< 0.05$ respectively). 
This indicates that storyboards and text may offer a more efficient means of information acquisition than video summaries which require sequential viewing. 
It is only with \Storyboard\ that the accuracy and time are both improved: with \Video\ and \Text, we observed a trade-off between accuracy and time. 

Recall that the summaries were generated live in the \Q\ condition. The model generation time was included in our measurement of task time. On average, to generate a summary  takes 
[$15.507, 9.426, 23.193$]${\second}$
with a standard deviation of [$0.246,0.298, 0.724,$]
for the modalities \Video, \Storyboard\ and \Text\ respectively. 
Generating the text summaries takes the longest time, as they require the video to be generated first. 
This is followed by the video summary, as the segments need to be concatenated to create a shorter video whereas the storyboard can be shown right after the selection process. 
While the query-based models were generating summaries, we observed that the users spent their time reading the questions or watching the long video. 
Even through it takes longer to generate \Text\ summaries, it is still more time efficient than \Video\ in total task time.

\subsection{Post-hoc Analysis}

\subsubsection{User Confidence}
While not directly related to the initial hypotheses, another variable of interest is the confidence of the users.  
Confidence was rated on a $1-10$ scale for object and event-related questions (sub-task level). 
However, some users displayed inconsistencies between their answers and confidence ratings: answering `Not sure' for all $3$ questions and giving a high confidence score indicating their high confidence in their uncertainty. 
To address this, the confidence score for a sub-task was assigned to each question under that sub-task, and `Not sure' answers were  excluded from the analysis. 
Since confidence was measured at a scale, it was modelled with ordinal logistic regression. 

The data was organised in the same way as the hypothesis testing. 
The results revealed a significant reduction in confidence when participants were provided with \G\ summaries compared to raw videos (coef. $\num{-0.4558}$, $p=0.011$). The drop in confidence for \Q\ was smaller than the drop in \G\ and was not significant (coef. $\num{-0.2119}$, $p=0.225>0.05$).  
This suggests a potential lack of trust in generic summaries where it was observed that the users' perception was that such systems might omit relevant information. 
User confidence in event questions were higher compared to object questions (coef. $\num{0.4396}$,  $p=0.002$), reiterating the findings in Section~\ref{subsubsub:first_accuracy} where participants had higher accuracy answering object questions.
No statistically significant relationships were found considering presentation modality.

\subsubsection{Usability}
Users gave a usability score per task answering the standard SUS questionnaire~\cite{brooke1996usability} 
which was modelled using an ordinal logistic regression model. 
We found no statistically significant relationships between the usability scores and the presence or absence of summary, or the type and modality of summary.

\subsubsection{Familiarity}
The pre-study survey consisted $8$ custom questions to capture the familiarity of the users with robots and video summarisation models, and their trust in such models. 
To analyse if there is a relationship between task performance (\score) and participant familiarity, a logistic regression model was fitted with individual questions forming the independent variables of the model. 
To keep the model simple, only the one-way interaction were considered. 
There were no statistical significance found between the score and familiarity, which suggests that the proposed video summary pipeline may lower the barrier of entry for novices. 

\subsubsection{Question-level analysis}
We further analysed the questions with the highest and lowest score amidst the $30$ participants. 
The lowest score for \Video\ was counting the blue barrels, where $16/30$ users said $1$, instead of the correct answer $2$.  
When users queried for a `blue barrel', the generated summary video often focused on the most prominent blue barrel, overlooking a second, less obvious barrel. \G\ summary was the same, only capturing the prominent blue barrel. This highlights a limitation in current models: an overemphasis on prominent features. 

Furthermore, the question with the lowest score for the \Storyboard\ task was `Did you knock down a door while trying to pass through it?'. 
Even though an image of the fallen door was included in the $24$ images provided under \G\ summary, $12/15$ users incorrectly answered `No' or `Not sure',  highlighting the fundamental drawback of the \Storyboard\ modality for event questions.

\subsubsection{Post-study survey answers}
During the post-study survey, the users were asked if the provided summaries were helpful in their task completion. 
Participants in the \Q\ group unanimously reported that AI tools helped them, compared to $11/15$ in the \G\ group. This 
highlighted the lack of information in generic summaries. 

The participants were also asked about their least and most preferred modality. 
To calculate a single preference score, the most preferred modality is given a score of \num{+1} and the least preferred \num{-1}, and the scores are averaged over each modality.  
The users' preference over the modality is given in Fig.~\ref{fig:posthoc} (\textit{left}). 
On average, users preferred \Video\ the most under the \G\ condition and equally \Video\ and \Storyboard\ under the \Q\ condition. 
Users disliked \Text\ with both summary types.
Interestingly, the \Storyboard\  showed a dramatic increase from the \G\ to \Q\ condition, implying image outputs are more useful in an interactive setting. 
The common reason stated by users for the preference for \Video\ 
was provision of contextual evidence, resulting in increased confidence, improved explainability, and greater trustworthiness, 
whereas with \Text, participants reported the lack of evidence being the main reason of choosing it as the least preferred modality.

\begin{figure}
 \centering
 \includegraphics[width=0.5\textwidth]{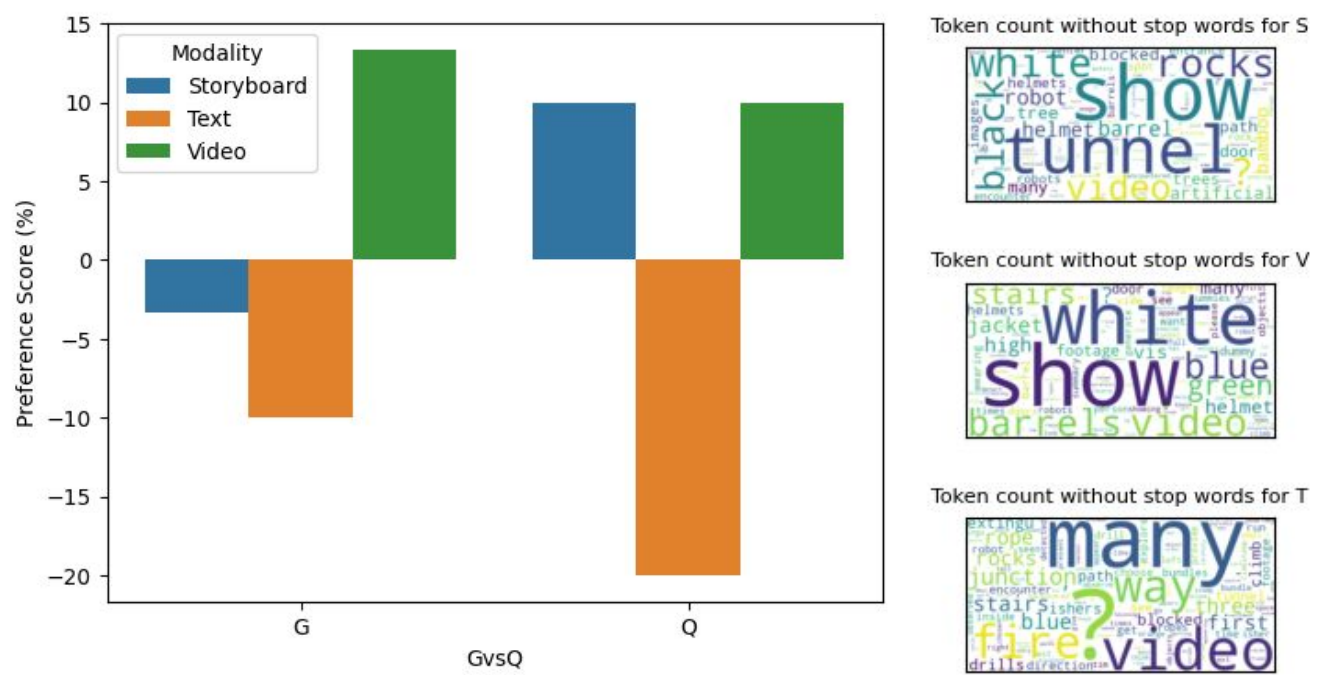}
 \caption{Post-hoc analysis. (\textit{left}): Preference score, (\textit{right}): Word clouds (token frequency) of user queries.}
 \vspace{-5mm}
 \label{fig:posthoc}
\end{figure}

\subsubsection{User Query Analysis}

User queries under each modality were broken down into tokens and plotted as word clouds (excluding common English stop words): the bigger a word, the more frequent it is (Fig.~\ref{fig:posthoc} \textit{right}). 
With modalities: \Video\ and \Storyboard, users more often make queries such as `show \dots', whereas with \Text\ the dominant style is a complete question: `how many \dots?'
With \Text, the average query length of \num{9.3} {words} was \qty{70}{\percent} higher than \Video\ and \Storyboard\ (\numlist{5.6;5.3} {words} respectively). 
Users made \qty{30}{\percent} more queries per video in \Storyboard\ than the other two modalities, however this may be due to the \Storyboard\ model being faster to calculate/load than the other models.
We also observed that the users had more difficulties forming queries to retrieve event-related information compared to object-related information in modalities \Video\ and \Storyboard. 

\subsection{Main Findings}
The study demonstrates that query-driven summaries significantly improve retrieval accuracy compared to raw data; however, this advantage comes with the trade-off of increased time spent on the task.  
Among the modalities, storyboards are particularly effective, enhancing both accuracy and reducing time compared to video and text summaries. 
However, storyboards are less effective for event-related questions, where other modalities might be more suitable. 
The preference for modality appears to be subjective, with storyboards and video equally favored on average in query-driven summaries, while text is least preferred. 
The study also suggests that utilizing ViFMs for intermittent supervision could potentially enable novices to perform at equivalent accuracy as more experienced users.  Our work demonstrates ViFM as a promising tool for intermittent robot supervision via query-driven summaries. 
Further, our findings on participants' performance difference in object vs. event questions highlight the room for improvement in ViFMs' capabilities in object localization and counting. 
} %

\section{CONCLUSIONS}
\label{sec:conclusion}

We investigated the efficacy of ViFMs in generating robot summaries for intermittent supervision. 
Our findings reveal both the potential and limitations of these models in distilling complex robot behavior into comprehensible summaries. 
Query-driven summaries significantly improved task accuracy compared to generic summaries and raw video, albeit at the cost of increased task completion time. 
The modality of summary presentation also played a crucial role, with storyboards potentially offering a balance between accuracy and efficiency compared to video and text formats.
The lack of correlation between user familiarity and  performance suggests that well-designed summaries can effectively bridge knowledge gaps for users with varying levels of expertise. 

Our work demonstrates the potential of ViFMs for improving robot-to-human and two-way communication. It serves as the foundation for future work to develop more sophisticated ViFMs which consider both linguistic subtleties and users' implicit knowledge structures when processing queries and adapt the summary presentation base on the task context.

\addtolength{\textheight}{-12cm}   %

\section*{ACKNOWLEDGMENT}
We thank Pavan Sikka, CSIRO, for his valuable insights and contributions to this work.

\bibliographystyle{IEEEtran}
\bibliography{example} %

\end{document}